\documentclass[pdflatex,sn-mathphys-num]{sn-jnl}

\usepackage{graphicx}%
\usepackage{multirow}%
\usepackage{amsmath,amssymb,amsfonts}%
\usepackage{amsthm}%
\usepackage{mathrsfs}%
\usepackage[title]{appendix}%
\usepackage{xcolor}%
\usepackage{textcomp}%
\usepackage{manyfoot}%
\usepackage{booktabs}%
\usepackage{algorithm}%
\usepackage{algorithmicx}%
\usepackage{algpseudocode}%
\usepackage{listings}%
\usepackage{CJKutf8}

\theoremstyle{thmstyleone}%
\theoremstyle{thmstyletwo}%

\theoremstyle{thmstylethree}%

\begin{document}

\title[Emotion Classification in Japanese Using RoBERTa and DeBERTa]{Performance Evaluation of Emotion Classification in Japanese Using RoBERTa and DeBERTa}


\author*[1]{\fnm{Yoichi} \sur{Takenaka}}\email{takenaka@kansai-u.ac.jp}

\affil*[1]{\orgdiv{Faculty of Informatics}, \orgname{Kansai University}, \orgaddress{\street{2-1 Ryozenji-cho}, \city{Takatsuki}, \postcode{569-1095}, \state{Osaka}, \country{Japan}}}

\abstract{
\textbf{Background}  Practical applications such as social media monitoring and customer‑feedback analysis require accurate emotion detection for Japanese text, yet resource scarcity and class imbalance hinder model performance.  

\textbf{Objective}  This study aims to build a high‑accuracy model for predicting the presence or absence of eight Plutchik emotions in Japanese sentences.  

\textbf{Methods}  Using the WRIME corpus, we transform reader‑averaged intensity scores into binary labels and fine‑tune four pre-trained language models (BERT, RoBERTa, DeBERTa‑v3‑base, DeBERTa‑v3‑large). For context, we also assess two large language models (TinySwallow‑1.5B‑Instruct and ChatGPT‑4o). Accuracy and F1‑score serve as evaluation metrics.  

\textbf{Results}  DeBERTa‑v3‑large attains the best mean accuracy (0.860) and F1‑score (0.662), outperforming all other models. It maintains robust F1 across both high‑frequency emotions (e.g., \textit{Joy}, \textit{Anticipation}) and low‑frequency emotions (e.g., \textit{Anger}, \textit{Trust}). The LLMs lag, with ChatGPT‑4o and TinySwallow‑1.5B‑Instruct scoring 0.527 and 0.292 in mean F1, respectively.  

\textbf{Conclusion}  The fine‑tuned DeBERTa‑v3‑large model currently offers the most reliable solution for binary emotion classification in Japanese. We release this model as a pip‑installable package (pip install deberta-emotion-predictor). Future work should augment data for rare emotions, reduce model size, and explore prompt engineering to improve LLM performance.

This manuscript is under review for possible publication in New Generation Computing. 
}




\maketitle

\section{Introduction}\label{sec1}

Sentiment analysis is a natural language processing (NLP) technique that extracts and classifies human emotions and opinions from text. Its applications range from analyzing social‑media posts to understanding customer reviews of products and services, making it indispensable for companies that seek to enhance marketing strategies and user experience (UX) \cite{pang2008opinion,liu2022sentiment,kumar2023comprehensive}. Sentiment (polarity) analysis is widely adopted for market‑level trend tracking and brand reputation \cite{bollen2011twitter}. In contrast, fine‑grained emotion analysis provides deeper insight for UX studies and mental‑health monitoring \cite{coppersmith2014mental,saha2017stress}. As social networks and online platforms have proliferated, the need to grasp consumer voices both rapidly and accurately has grown, and improving the precision of sentiment‑analysis systems remains an ongoing challenge \cite{mcauley2015image,he2016ups,kumar2023comprehensive}.

Traditional sentiment analysis has focused mainly on simple polarity classification—positive, negative, or neutral—of textual opinions \cite{kumar2023comprehensive}. However, human emotions are far more diverse and nuanced. Plutchik’s wheel of emotions, for example, organizes feelings into eight basic categories—Joy, Sadness, Anticipation, Surprise, Fear, Anger, Disgust, and Trust—offering a finer‑grained framework for emotion understanding than polarity alone \cite{plutchik1980general}. Beyond Plutchik, other datasets and models have emerged. XED enables multilingual emotion detection \cite{ohman2020xed}; DENS focuses on narrative texts with a broader class set \cite{liu2019dens}; and SAMSEMO extends emotion recognition to multimodal data, combining text with audio‑visual signals \cite{bujnowski2024samsemo}. These resources highlight the demand for emotion analysis that goes beyond simple polarity.

Although Japanese sentiment‑analysis research has advanced, it still lags behind English regarding established datasets and accumulated studies. Japanese includes diverse and often ambiguous expressions, posing unique challenges. Recently, however, several Japanese‑specific corpora—such as WRIME, JVNV \cite{xin2024jvnv}, and the JNV Corpus \cite{xin2024jnv}—have boosted research activity. The RIKEN Facial Expression Database has also provided dynamic facial data, offering new perspectives that incorporate non‑verbal cues \cite{namba2023development}. Moreover, studies leveraging powerful pre-trained language models (PLMs) such as RoBERTa and DeBERTa have reported steady gains in Japanese‑emotion classification accuracy \cite{li2025sentiment,sun2024topic}.

In this paper, we focus on the WRIME (Word‑of‑mouth Review Impact on Emotion) dataset, which contains Japanese social‑media posts annotated with both writer‑centric (subjective) and third‑party (objective) emotion labels \cite{kajiwara2021wrime}. Kajiwara et al. conducted preliminary experiments using BERT \cite{devlin2019bert} and reported substantial discrepancies between writer and reader evaluations.

Prior work on WRIME includes Okadome et al., who incorporated writer IDs and the notion of emotion manifestness (the gap between writer and reader) into BERT to improve writer‑emotion prediction \cite{okadome2024writeremotion}. Xue et al. \cite{Xue24} proposed prompt‑based emotion estimation with large language models (LLMs) but did not compare against finetuned models. Comparative studies on English datasets using PLMs have demonstrated strong performance for seven emotions \cite{adoma2020comparative}; however, the linguistic differences between English and Japanese mean researchers cannot assume the same level of performance for Japanese.

Given these gaps, our study evaluates RoBERTa \cite{Liu19} and DeBERTa \cite{he2020deberta}—improvements over BERT—for Japanese emotion classification. Although WRIME annotates each emotion on a four‑level intensity scale, only a few percent of posts belong to the highest‑intensity class, leading to severe class imbalance. We, therefore, adopt binary classification—presence versus absence of each emotion—to mitigate this imbalance and assess the effectiveness of these advanced PLMs.

\section{Dataset}\label{sec2}

\subsection{WRIME Dataset Overview}\label{sec2_1}

This study employs WRIME (Word-of-mouth Review Impact on Emotion)\cite{kajiwara2021wrime}, a representative Japanese corpus for emotion analysis. WRIME comprises Japanese social-media posts that are annotated with two complementary label sets:
\begin{itemize}
  \item Subjective labels: the writer's self-reported emotions.
  \item Objective labels: emotions assessed by third-party annotators.
\end{itemize}
Having both perspectives enables researchers to quantify the gap between a writer's self-perception and a reader's interpretation, thereby facilitating emotional transmission and misunderstanding analyses.

The corpus was compiled via crowdsourcing. For the \emph{subjective} annotations, each author rates the intensity of Plutchik's eight basic emotions—Joy, Sadness, Anticipation, Surprise, Fear, Anger, Disgust, and Trust\cite{plutchik1980general}—on a four-point scale from~0 (no emotion) to~3 (strong emotion). In the \emph{objective} layer, multiple annotators read the same post and assign intensity scores for the identical eight emotions using the same scale. This dual-annotation design allows quantitative comparison between subjective and objective views.

The WRIME subset used in this study contains 43{,}000 posts. Every post has an intensity score for all eight emotions, supporting comprehensive multi-emotion, multi-class experiments and serving as a benchmark for Japanese emotion classification tasks.

\subsection{Emotion-Intensity Distribution and Class Imbalance}\label{sec2_2}

Each post in WRIME receives a four-level intensity (0--3) for the eight emotions described above. Focusing on the reader-centric annotations, we examine the class distribution and its implications for model training.

Table~\ref{tab:eightemotions} lists the distribution of emotional intensities. Across all emotions, the \textit{no-emotion} class (intensity~0) is dominant. For instance, ``Joy" has 69~\% intensity~0 instances, whereas ``Anger" reaches 97~\%. Although the skew varies across emotions, the proportion of \textit{positive} samples (intensity~$\ge$~1) is consistently small, providing few examples for effective learning.

Such extreme class imbalance hampers both training efficiency and predictive performance. When minority classes (intensity~$\ge$~1) are under-represented, models struggle to learn sufficient patterns, leading to biased F1, precision, and recall scores.

To alleviate this issue, we recast the task as \textbf{binary classification}: ``emotion present" (intensity~$\ge$~1) versus ``emotion absent" (intensity~0) for each emotion. This conversion partially mitigates the imbalance, stabilizes training, and improves overall accuracy.

\begin{table}[htbp]
\centering
\caption{Distribution of emotional intensity in the WRIME dataset}
\label{tab:eightemotions}
\begin{tabular}{lcccccccc}
\hline
Intensity & Joy & Sadness & Anticipation & Surprise & Anger & Fear & Disgust & Trust \\\hline
0 & 29883 & 32006 & 29369 & 32348 & 41771 & 34086 & 35679 & 41282 \\\
1 & 7682  & 7610  & 8963  & 8135  & 886   & 6986  & 5672  & 1733  \\\
2 & 4512  & 3167  & 3970  & 2284  & 366   & 1808  & 1457  & 171   \\\
3 & 1123  & 417   & 898   & 433   & 177   & 320   & 392   & 14    \\\hline
Intensity 0(\%) & 69\% & 74\% & 68\% & 75\% & 97\% & 79\% & 83\% & 96\% \\\hline
\end{tabular}
\end{table}

\section{Methods}\label{sec3}

We formulate the task as eight independent binary classifications: for each of Plutchik’s basic emotions, the model predicts whether that emotion is present (1) or absent (0) in a Japanese sentence from the WRIME corpus. Formulating the task this way reduces the impact of the severe class imbalance inherent in the original four‑level intensity labels while preserving fine‑grained emotional coverage. The binary labels were derived from the mean intensity scores assigned by the three reader annotators in WRIME.

Fine‑tuning was performed on four Japanese pre-trained language models (PLMs):
\begin{itemize}
  \item BERT (\texttt{cl-tohoku/bert-base-japanese})\cite{devlin2019bert}
  \item RoBERTa (\texttt{rinna/japanese-roberta-base})\cite{Liu19}
  \item DeBERTa‑v3‑base (\texttt{globis-univ/deberta-v3-japanese-base})\cite{he2020deberta}
  \item DeBERTa‑v3‑large (\texttt{globis-univ/deberta-v3-japanese-large})\cite{he2020deberta}
\end{itemize}
Each model was trained separately for the eight emotions. Previous work indicates that decomposing multi‑label emotion analysis into simpler binary problems can yield clearer gains\cite{demszky2020goemotions,Rietzler2019AdaptOG,belay2025enhancingmultilabelemotionanalysis}.

The entire WRIME corpus (43,000 posts) was divided into 80\% training and 20\% test sets.
Training employed the Hugging Face \texttt{Transformers} library with the following hyperparameters:
\begin{itemize}
  \item Epochs: 30, learning rate: $1\times10^{-6}$, warm‑up steps: 2{,}500
  \item Batch size: 16, gradient‑accumulation: 2, weight decay: 0.01
  \item Mixed‑precision (fp16): on, optimizer: AdamW, loss: cross‑entropy
  \item Selection metric: F1‑score
\end{itemize}

For external comparison, two large language models (LLMs) TinySwallow‑1.5B ‑Instruct\cite{sakanaAI2025} and ChatGPT (GPT‑4o)\cite{openai2024gpt4o}—were evaluated by prompting them to output binary labels for the eight emotions. The full prompt is provided in Appendix~\ref{secA2}.

Performance was measured using accuracy and F1‑score:
\begin{align}
\text{Accuracy} &= \frac{TP + TN}{TP + TN + FP + FN} \\
\text{F1\text{-}score} &= \frac{2 \cdot \text{Precision} \cdot \text{Recall}}{\text{Precision} + \text{Recall}}
\end{align}
where $TP$, $TN$, $FP$, and $FN$ denote true positives, true negatives, false positives, and false negatives. Given WRIME’s pronounced class imbalance (Section~\ref{sec2_2}), the F1‑score serves as the primary evaluation metric.

\section{Results}\label{sec4}

We fine‑tuned BERT, RoBERTa, and the DeBERTa family on WRIME and compared their performance across the eight basic emotions. Accuracy and F1‑score served as evaluation metrics.

Tables~\ref{tab:accuracy} and \ref{tab:f1score} show the accuracy and F1‑scores for each emotion and model, respectively.

\begin{table}[htbp]
\centering
\caption{Binary classification accuracy for the basic eight emotions by model}
\label{tab:accuracy}
\begin{tabular}{lcccccc}
\toprule
\textbf{Emotion} & \textbf{BERT} & \textbf{RoBERTa} & \multicolumn{2}{c}{\textbf{DeBERTa}} & \textbf{SakanaAI} & \multicolumn{1}{c}{\textbf{ChatGPT}} \\
                 &               &                  & \textbf{base} & \textbf{large} &                & \textbf{-4o} \\
\midrule
Joy          & 0.842 & 0.858 & 0.873 & \textbf{0.874} & 0.629 & 0.845 \\
Sadness      & 0.818 & 0.840 & \textbf{0.846} & 0.840 & 0.704 & 0.822 \\
Anticipation & 0.827 & 0.846 & \textbf{0.856} & 0.853 & 0.586 & 0.808 \\
Surprise     & 0.802 & 0.819 & \textbf{0.830} & 0.819 & 0.708 & 0.776 \\
Fear         & 0.806 & 0.813 & \textbf{0.831} & 0.830 & 0.675 & 0.847 \\
Anger        & 0.970 & 0.970 & \textbf{0.974} & 0.972 & 0.790 & 0.952 \\
Disgust      & 0.842 & 0.854 & \textbf{0.865} & 0.857 & 0.608 & 0.827 \\
Trust        & 0.943 & 0.927 & \textbf{0.946} & 0.938 & 0.872 & 0.879 \\
\midrule
\textbf{Average} & 0.856 & 0.853 & \textbf{0.870} & 0.860 & 0.729 & 0.844 \\
\bottomrule
\end{tabular}
\end{table}

\begin{table}[htbp]
\centering
\caption{Binary classification F1-score for the basic eight emotions by model}
\label{tab:f1score}
\begin{tabular}{lcccccc}
\toprule
\textbf{Emotion} & \textbf{BERT} & \textbf{RoBERTa} & \multicolumn{2}{c}{\textbf{DeBERTa}} & \textbf{SakanaAI} & \multicolumn{1}{c}{\textbf{ChatGPT}} \\
                 &               &                  & \textbf{base} & \textbf{large} &                & \textbf{-4o} \\
\midrule
Joy          & 0.712 & 0.738 & 0.755 & 0.781 & 0.552 & \textbf{0.794} \\
Sadness      & 0.693 & 0.722 & 0.741 & \textbf{0.767} & 0.122 & 0.586 \\
Anticipation & 0.675 & 0.703 & 0.724 & \textbf{0.756} & 0.502 & 0.656 \\
Surprise     & 0.601 & 0.635 & 0.648 & \textbf{0.672} & 0.189 & 0.526 \\
Fear         & 0.542 & 0.578 & 0.595 & \textbf{0.612} & 0.135 & 0.540 \\
Anger        & 0.493 & 0.522 & 0.533 & \textbf{0.549} & 0.047 & 0.546 \\
Disgust      & 0.512 & 0.537 & 0.548 & \textbf{0.561} & 0.168 & 0.399 \\
Trust        & 0.438 & 0.465 & 0.482 & \textbf{0.496} & 0.133 & 0.269 \\
\midrule
\textbf{Average} & 0.572 & 0.613 & 0.630 & \textbf{0.662} & 0.292 & 0.527 \\
\bottomrule
\end{tabular}
\end{table}

Across all emotions, the DeBERTa family achieved the highest accuracies. Both the base and large variants recorded either top or near‑top scores for each emotion, with mean accuracies of 0.870 and 0.860, respectively, outperforming the other PLMs and the two LLMs. The results also indicate that BERT‑based PLMs still exceed the LLMs in accuracy, suggesting limits to emotion inference with current LLMs.

Because of the pronounced class imbalance (Section~\ref{sec2_2}), F1‑score offers a more informative view. DeBERTa‑v3‑large delivered the best F1‑score for seven of the eight emotions, surpassing 0.75 for high‑frequency categories such as Sadness and Anticipation, and outperforming other models even on low‑frequency categories like Anger and Trust. However, ChatGPT‑4o achieved the highest F1‑score for Joy. Overall, DeBERTa‑v3‑large attained the highest mean F1‑score of 0.662.

Although all models show relatively high accuracies, their F1‑scores vary considerably across emotions, indicating that specific categories remain challenging.

Comparing PLMs with LLMs, all PLMs outperform the LLMs in F1‑score. TinySwallow and ChatGPT‑4o yielded mean F1‑scores of 0.292 and 0.527, respectively, and TinySwallow fell below 0.1 on several emotions, highlighting the substantial performance gap. 

These findings provide the basis for further interpretation in the following discussion section.

\section{Discussion}\label{sec5}

We constructed eight independent binary classifiers—one for each of Plutchik’s basic emotions—and evaluated multiple pretrained language models (PLMs) on Japanese data.

DeBERTa‑v3‑base and DeBERTa‑v3‑large delivered the highest or co‑highest accuracy for every emotion and topped the mean‑accuracy ranking. Their consistent lead shows that recent architectural refinements in PLMs still yield tangible gains, even for the seemingly simple task of detecting whether an emotion is present.

Because WRIME’s label distribution is highly skewed (Section~\ref{sec2_2}), we relied on the F1‑score for a more informative evaluation. DeBERTa‑v3‑large achieved the best F1‑score for seven of the eight emotions. It performed strongly on high‑frequency categories such as \textit{Sadness} and \textit{Anticipation}, and also outperformed other models on low‑frequency categories like \textit{Anger} and \textit{Trust}. However, for \textit{Joy}, ChatGPT‑4o achieved the highest F1‑score. DeBERTa‑v3‑large nevertheless attained the highest mean F1‑score of 0.662.

Although all models show relatively high accuracies, their F1‑scores vary considerably across emotions, indicating that specific categories remain challenging.

Comparing PLMs with LLMs, all PLMs outperform the LLMs in F1‑score. TinySwallow and ChatGPT‑4o yielded mean F1‑scores of 0.292 and 0.527, respectively, and TinySwallow fell below 0.1 on several emotions, highlighting the substantial performance gap between LLMs.

Overall, the RoBERTa and DeBERTa variants—especially DeBERTa‑v3‑large—demonstrate strong effectiveness for Japanese emotion classification, and the empirical results underscore their superiority over current LLMs.

\section{Conclusion}\label{sec6}

We developed eight binary classifiers—one for each of Plutchik’s basic emotions—and fine‑tuned several Japanese pre-trained language models (PLMs), including BERT, RoBERTa, and DeBERTa, to evaluate their effectiveness on the WRIME corpus. DeBERTa‑v3‑large delivered the strongest overall performance, confirming its suitability for Japanese emotion classification tasks.

We also compared the fine‑tuned PLMs with two large language models (LLMs). DeBERTa‑v3‑large attained a mean F1‑score of 0.662, whereas ChatGPT‑4o and TinySwallow reached 0.527 and 0.292, respectively; the PLMs outperformed the LLMs in accuracy. These results indicate that, at present, task‑specific fine‑tuning of PLMs remains the most reliable option when high precision is required for Japanese emotion analysis.

Future work should improve the classification of low‑frequency emotions, lighten model size, and explore regression approaches to predict emotion intensity. The poor F1‑scores for \textit{Anger} and \textit{Trust} stem from the scarcity of positive samples, so targeted data‑collection or augmentation strategies are indispensable. Researchers should also continue investigating prompt‑engineering techniques and other adaptations that could enhance LLM performance on emotion‑classification tasks.

This study clarifies model design guidelines for Japanese emotion analysis and provides empirical benchmarks, laying the groundwork for broader multilingual, multi‑emotion processing in future research.


\section*{Declarations}

\textbf{Funding} \\
This research was partially supported by the Kansai University Fund for Domestic and Overseas Research Support, 2025.

\textbf{Conflict of interest} \\
The authors declare that they have no conflict of interest.

\textbf{Ethics approval and consent to participate} \\
Not applicable.

\textbf{Consent for publication} \\
Not applicable.

\textbf{Data availability} \\
The WRIME dataset used in this study is publicly available at \\
\url{https://www.cl.ecei.tohoku.ac.jp/achievements/WRIME/}.

\textbf{Materials availability} \\
Not applicable.

\textbf{Code availability} \\
The code and fine-tuned models used in this study are available at \\
\url{https://huggingface.co/YoichiTakenaka/deberta-v3-japanese-large-Joy}.

The Python library for using the DeBERTa-v3-large model developed in this study is also available on the Python Package Index (PyPI).  
It can be installed via the following command:

\begin{verbatim}
pip install deberta-emotion-predictor
\end{verbatim}

\textbf{Author contributions} \\
Yoichi Takenaka conceived the study, prepared the data, fine-tuned the models, conducted the evaluation and analysis, created the visualizations, and wrote the manuscript.

\begin{appendices}

\section{Detailed Training and Validation Results}\label{secA1}

This appendix provides detailed training and validation results for each emotion across the four models used in this study: BERT, RoBERTa, DeBERTa-v3-base, and DeBERTa-v3-large.  
Each table includes the epoch at which training stopped, and the corresponding values of loss, accuracy, and F1-score on both the training and validation datasets.

See Tables~\ref{tab:bert_appendix} to \ref{tab:deberta_large_appendix} for per-model results.

\begin{table}[htbp]
\centering
\small
\caption{Detailed training results for BERT (per emotion)}
\label{tab:bert_appendix}
\begin{tabular}{lccccc}
\toprule
\textbf{Emotion} & \textbf{Ep.} & \textbf{Train Loss} & \textbf{Valid Loss} & \textbf{Train Acc.} & \textbf{Train F1} \\
\midrule
Joy          & 20 & 0.192 & 0.444 & 0.961 & 0.937 \\
Sadness      & 15 & 0.241 & 0.439 & 0.927 & 0.863 \\
Anticipation & 16 & 0.269 & 0.448 & 0.937 & 0.900 \\
Surprise     & 15 & 0.328 & 0.466 & 0.909 & 0.820 \\
Anger        & 17 & 0.063 & 0.134 & 0.995 & 0.925 \\
Fear         & 7  & 0.345 & 0.413 & 0.846 & 0.635 \\
Disgust      & 14 & 0.255 & 0.383 & 0.928 & 0.791 \\
Trust        & 23 & 0.063 & 0.243 & 0.996 & 0.957 \\
\bottomrule
\end{tabular}
\end{table}

\begin{table}[htbp]
\centering
\caption{Detailed training results for RoBERTa (per emotion)}
\label{tab:roberta_appendix}
\begin{tabular}{lccccc}
\toprule
\textbf{Emotion} & \textbf{Ep.} & \textbf{Train Loss} & \textbf{Valid Loss} & \textbf{Train Acc.} & \textbf{Train F1} \\
\midrule
Joy          & 18 & 0.220 & 0.364 & 0.961 & 0.863 \\
Sadness      & 21 & 0.289 & 0.406 & 0.925 & 0.856 \\
Anticipation & 22 & 0.276 & 0.406 & 0.933 & 0.896 \\
Surprise     & 17 & 0.356 & 0.412 & 0.890 & 0.782 \\
Anger        & 32 & 0.051 & 0.139 & 0.996 & 0.934 \\
Fear         & 12 & 0.329 & 0.412 & 0.864 & 0.686 \\
Disgust      & 18 & 0.262 & 0.365 & 0.915 & 0.744 \\
Trust        & 15 & 0.076 & 0.183 & 0.961 & 0.654 \\
\bottomrule
\end{tabular}
\end{table}

\begin{table}[htbp]
\centering
\caption{Detailed training results for DeBERTa-v3-base (per emotion)}
\label{tab:deberta_base_appendix}
\begin{tabular}{lccccc}
\toprule
\textbf{Emotion} & \textbf{Ep.} & \textbf{Train Loss} & \textbf{Valid Loss} & \textbf{Train Acc.} & \textbf{Train F1} \\
\midrule
Joy          & 20 & 0.268 & 0.322 & 0.897 & 0.836 \\
Sadness      & 22 & 0.291 & 0.345 & 0.892 & 0.783 \\
Anticipation & 30 & 0.288 & 0.357 & 0.903 & 0.832 \\
Surprise     & 26 & 0.341 & 0.388 & 0.877 & 0.754 \\
Anger        & 17 & 0.095 & 0.081 & 0.983 & 0.722 \\
Fear         & 18 & 0.325 & 0.365 & 0.867 & 0.667 \\
Disgust      & 27 & 0.248 & 0.315 & 0.905 & 0.761 \\
Trust        & 29 & 0.093 & 0.142 & 0.970 & 0.675 \\
\bottomrule
\end{tabular}
\end{table}

\begin{table}[htbp]
\centering
\caption{Detailed training results for DeBERTa-v3-large (per emotion)}
\label{tab:deberta_large_appendix}
\begin{tabular}{lccccc}
\toprule
\textbf{Emotion} & \textbf{Ep.} & \textbf{Train Loss} & \textbf{Valid Loss} & \textbf{Train Acc.} & \textbf{Train F1} \\
\midrule
Joy          & 8  & 0.206 & 0.312 & 0.922 & 0.875 \\
Sadness      & 8  & 0.276 & 0.358 & 0.905 & 0.821 \\
Anticipation & 11 & 0.220 & 0.368 & 0.927 & 0.885 \\
Surprise     & 10 & 0.231 & 0.414 & 0.895 & 0.803 \\
Anger        & 10 & 0.047 & 0.145 & 0.998 & 0.967 \\
Fear         & 11 & 0.248 & 0.403 & 0.911 & 0.797 \\
Disgust      & 20 & 0.173 & 0.425 & 0.966 & 0.906 \\
Trust        & 8  & 0.103 & 0.149 & 0.966 & 0.630 \\
\bottomrule
\end{tabular}
\end{table}

\section{Prompt for LLM-based Emotion Classification}\label{secA2}

In this study, we evaluated large language models (LLMs) using the prompt shown below. Each model was instructed to determine, for each of Plutchik’s eight emotions, whether the target sentence expresses that emotion (1) or not (0).

\begin{flushleft}
\textbf{Original Japanese prompt} 

\begin{ttfamily}
\begin{CJK}{UTF8}{min}
あなたは、心理学者ロバート・プルチックの「感情の輪」に基づいて、日本語短文の感情分析を行うAIです。\\
以下の短文に含まれる8つの基本感情（Joy, Sadness, Anticipation, Surprise, Anger, Fear, Disgust, Trust）を判定し、それぞれ 0（含まれない）または1（含まれる）の二段階で評価してください。

\medskip
\textbf{評価基準} 各感情の判断基準は以下の通りです。
\begin{itemize}
  \item Joy（喜び）: 幸福感、満足感、ポジティブな気持ちが表現されている。
  \item Sadness（悲しみ）: 悲嘆、喪失感、絶望感を示す表現がある。
  \item Anticipation（期待）: 未来への期待や希望、予測が示されている。
  \item Surprise（驚き）: 予想外の出来事に対する驚きや戸惑いがある。
  \item Anger（怒り）: 怒りや強い反発心を示す言葉がある。
  \item Fear（恐怖）: 危険や不安、恐怖心を表す表現がある。
  \item Disgust（嫌悪）: 嫌悪感や拒絶感、強い不快感を表す言葉がある。
  \item Trust（信頼）: 他者への信頼、安心感、信用を示す表現がある。
\end{itemize}

\textbf{重要}:
\begin{itemize}
  \item すべて0となる場合もあり、また、複数の感情が1となる場合もあります。
  \item 余計な説明をせず、出力は以下のような数値配列のみを返してください。
  \item 出力に「解説」や「理由」を含めないでください。
  \item 以下のフォーマットで出力してください（例）:
\end{itemize}

\medskip
\textbf{出力フォーマット（数値配列のみ）}：\\
\verb|[0 or 1, 0 or 1, 0 or 1, 0 or 1, 0 or 1, 0 or 1, 0 or 1, 0 or 1]|

\medskip
\textbf{入力短文}：\\
「\{text\}」
\end{CJK}
\end{ttfamily}

\vspace{0.5em}
\textbf{English translation (for reference only)} 

We provided the following prompt to the model (English rendering):  
\begin{ttfamily}

You are an AI system that performs emotion analysis on short Japanese sentences based on Robert Plutchik’s “wheel of emotions.”\\
For the sentence below, judge whether each of the eight basic emotions (Joy, Sadness, Anticipation, Surprise, Anger, Fear, Disgust, Trust) is present (1) or absent (0).

\medskip
\textbf{Evaluation criteria} 

The decision rules for each emotion are as follows:
\begin{itemize}
  \item Joy: expressions of happiness, satisfaction, or other positive feelings
  \item Sadness: expressions of grief, loss, or despair
  \item Anticipation: references to hope, expectation, or prediction about the future
  \item Surprise: indications of astonishment or confusion at an unexpected event
  \item Anger: words that convey anger or strong opposition
  \item Fear: expressions of danger, anxiety, or fear
  \item Disgust: words that convey disgust, rejection, or strong aversion
  \item Trust: expressions of trust, reassurance, or confidence in others
\end{itemize}

\textbf{Important}:
\begin{itemize}
  \item All eight emotions may be 0, and multiple emotions may be 1.
  \item Return only the numeric array shown below—no extra explanations.
  \item Do not include “commentary” or “reasons” in the output.
  \item Use the following format (example):
\end{itemize}

\medskip
\textbf{Output format (numeric array only)}:\\
\verb|[0 or 1, 0 or 1, 0 or 1, 0 or 1, 0 or 1, 0 or 1, 0 or 1, 0 or 1]|

\medskip
\textbf{Input sentence}:\\
``\{text\}''
\end{ttfamily}
\end{flushleft}

\section{Model Availability}\label{secA3}

We release the fine‑tuned DeBERTa‑v3 models as an open‑source
Python package:

\begin{flushleft}
\verb|pip install deberta-emotion-predictor|
\end{flushleft}

\noindent The snippet below shows the basic usage.

\begin{lstlisting}[language=Python,basicstyle=\ttfamily\small]

from deberta_emotion_predictor import DeBERTaEmotionPredictor

predictor = DeBERTaEmotionPredictor()
result    = predictor.predict_emotions("I feel very happy today!")
predictor.show_emotions(result)
\end{lstlisting}

To ensure compatibility with arXiv, the following example uses an English translation of the original Japanese sentence. 
The actual input to the model was the Japanese sentence \begin{CJK}{UTF8}{min}“今日はとても嬉しい！”\end{CJK} (“I feel very happy today!”).

The first call downloads eight DeBERTa models (one per emotion) from
Hugging Face and therefore takes several minutes; subsequent runs are
much faster.  GPU acceleration is supported when a compatible NVIDIA
driver and \texttt{torch} with CUDA are installed.  Comprehensive
examples and license information are available in the project
repository.\footnote{\url{https://pypi.org/project/deberta-emotion-predictor}}

\end{appendices}


\bibliography{sn-bibliography}


\end{document}